\documentclass[letterpaper, 10 pt, conference]{ieeeconf}  

\IEEEoverridecommandlockouts                              

\usepackage{amsfonts}
\usepackage{algorithm}
\usepackage[algo2e, ruled]{algorithm2e}
\usepackage[dvipsnames]{xcolor}

\usepackage{physics}
\usepackage{siunitx}
\usepackage{textcomp}
\usepackage{comment}
\usepackage[hidelinks]{hyperref}
\usepackage{times}
\usepackage[symbol]{footmisc}
\usepackage{array}
\usepackage{multicol}
\usepackage{amsmath}
\usepackage{graphicx}
\usepackage{cleveref}
\usepackage{tikz}
\usepackage{makecell}
\usepackage{cite}
\usepackage{booktabs}

\usepackage{titlesec}

\titlespacing{\section}{0pt}{*0.5}{*0.5} 
\titlespacing{\subsection}{0pt}{*0.5}{*0.5} 
\titlespacing{\subsubsection}{0pt}{*0.5}{*0.5} 

\AtBeginDocument{\RenewCommandCopy\qty\SI}

\DeclareMathOperator*{\argmin}{arg\,min}

\newcolumntype{L}[1]{>{\raggedright\let\newline\\\arraybackslash\hspace{0pt}}m{#1}}
\newcolumntype{C}[1]{>{\centering\let\newline\\\arraybackslash\hspace{0pt}}m{#1}}
\newcolumntype{R}[1]{>{\raggedleft\let\newline\\\arraybackslash\hspace{0pt}}m{#1}}

\def\horizon{5}
\def\nsteps{25}
\def\lda{$10^{-3}$}

\def\lmax{6160}
\def\lscale{600}

\newcommand{\transp}{\mathsf{T}}
\newcommand{\R}{\mathbb{R}}
\newcommand\scalemath[2]{\scalebox{#1}{\mbox{\ensuremath{\displaystyle #2}}}}

\newcommand{\legendline}[2]{%
  \tikz[baseline=-0.6ex]{\draw[line width=1.5pt, line cap=round, #1] (0,0) -- (0.4,0);}~#2%
}

\newcommand{\legenddoublecircle}[3]{%
  \tikz[baseline=-0.6ex]{
    \node[circle, draw=black, fill=black, minimum size=4.5pt, inner sep=0pt] (outer) {};
    \node[circle, draw=#1, fill=#1, minimum size=4pt, inner sep=0pt] at (outer.center) {};
    \node[circle, draw=#1, fill=#2, minimum size=2pt, inner sep=0pt] at (outer.center) {};
  }~#3%
}

\title{\LARGE \bf
Multi-Step Gaussian Process Propagation for Adaptive Path Planning 
}

\author{Alex Beaudin$^{*1}$, Bjørn Andreas Kristiansen$^{*2}$, Kristoffer Gryte$^{2}$, Corrado Chiatante$^{3}$, Morten Omholt Alver$^{2}$,  \\Murat Arcak$^{1}$, Tor Arne Johansen$^{2}$
\thanks{$^{*}$A. Beaudin and B. A. Kristiansen contributed equally to this article and should both be considered first authors.}%
\thanks{$^{1}$Alex Beaudin and Murat Arcak are with the Department of Electrical Engineering and Computer Sciences,
      University of California, Berkeley, CA 94720, USA
      E-mail A. Beaudin: {\tt\small a.b@berkeley.edu}}%
\thanks{$^{2}$Bjørn Andreas Kristiansen, Kristoffer Gryte, Morten Omholt Alver, and Tor Arne Johansen are with the Department of Engineering Cybernetics, Norwegian University of Science and Technology (NTNU), NO-7491, Norway. E-mail B. A. Kristiansen: {\tt\small bjorn.a.kristiansen@ntnu.no}}%
\thanks{$^{3}$Corrado Chiatante is with the Department of Electronic Systems, Norwegian University of Science and Technology (NTNU), NO-7491, Norway.}%
}

\begin{document}
\bstctlcite{IEEEexample:BSTcontrol} 

\maketitle
\thispagestyle{empty}
\pagestyle{empty}

\begin{abstract}
Efficient and robust path planning hinges on combining all accessible information sources.
In particular, the task of path planning for robotic environmental exploration and monitoring depends highly on the current belief of the world. 
To capture the uncertainty in the belief, we present a Gaussian process based path planning method that adapts to multi-modal environmental sensing data and incorporates state and input constraints.
To solve the path planning problem, we optimize over future waypoints in a receding horizon fashion, and our cost is thus a function of the Gaussian process posterior over all these waypoints. 
We demonstrate this method, dubbed OLAh-GP, on an autonomous surface vessel using oceanic algal bloom data from both a high-fidelity model and in-situ sensing data in a monitoring scenario.
Our simulated and experimental results demonstrate significant improvement over existing methods. 
With the same number of samples, our method generates more informative paths and achieves greater accuracy in identifying algal blooms in \textit{chlorophyll a} rich waters, measured with respect to total misclassification probability and binary misclassification rate over the domain of interest.
\end{abstract}

\section{Introduction}

Navigating complex and uncertain environments is a core challenge in robotics \cite{brunner2013, chai2021, chen2021}.
Robots must make decisions based on uncertain and partial information of their state and environment -- both of which may be evolving over the task horizon.
Additionally, in many practical settings, the measure of task progress is ambiguous or only partially observable, see \cite{kaelbling1998}.
This requires incorporating new information as it becomes available, as well as actively seeking out relevant information.
In particular, for the task of path planning, it is essential to find paths that maximize information gain while ensuring safe and reliable operation.

The Informative Path Planning (IPP) problem casts the path planning problem as an information gain problem.
IPP refers to the broad problem of generating trajectories based on information metrics.
Among the tools for reasoning about information and uncertainty, Gaussian processes (GPs) are powerful nonparametric tools for modeling the state or environment.
Thus, GPs have been used extensively in robotics literature for path planning \cite{xiao2022, binney2010, dutta2025, marchant2014}, motion planning \cite{mukadam2016}, and state estimation \cite{deisenroth2015}.
They can efficiently incorporate new data as it becomes available, and provide meaningful predictive uncertainty on the modeled variables.
These properties make GPs ideal candidates both as an environmental representation and as a basis for trajectory optimization costs.
They have been widely used to guide exploration by either modeling the environment~\cite{manjanna2017, tokekar2016} and selecting points to visit based on the uncertainty.
However, existing work relies on greedy variance-reduction algorithms to generate paths, or restrict their attention to discrete domains.
For example, \cite{ott2024, dutta2025} solve the IPP on a graph by casting it as a mixed integer program, limiting both scalability and resolution.

In \cite{manjannaHeterogeneousMultiRobotSystem2018}, the highest variance (nearby) point is picked to explore at each iteration.
In \cite{berget2022}, Berget et al. use a weighted combination of the current variance and mean of the GP and select the point in a discretized grid with the highest value.
A similar approach is proposed in \cite{ge2022} which plans paths into the future using ocean current data and RRT$^*$.
In a similar setting, Foss et al.~\cite{foss_using_2022} use single-step forecasting and linearize the dynamics,
and Manjanna et al.~\cite{manjanna2017} use value iteration to find the next action.
Tokekar et al.~\cite{tokekar2016} use the Gaussianity of the GP to derive a probability of misclassification to guide planning. 
Points that are likely to be misclassified are chosen and then a traveling salesperson problem (TSP) solver is used to determine the shortest path.
Xiao et al.~\cite{xiao2022} find $n$ points to explore by selecting the highest variance point sequentially $n$ times and finding the shortest path visiting them all using a TSP solver.
Despite the wide variety of existing methods in continuous space, all plan myopically.
Although they select the next $n$ points, they forecast the GP uncertainty change at most one step at a time.
Thus, they omit the effects of samples in the planning horizon on the GP.
More recently, \cite{swindell2026} analyze the effects of discretization using an exhaustive planner.
They enumerates all paths of length four in a grid and selects the one which maximizes mutual information. The short horizon makes it comparable to a greedy approach.

Dual to IPP, the Bayesian sampling framework is a broader category of methods which sample based on information gain.
These methods balance exploration and exploitation under uncertainty, and borrow Bayesian inference tools.
Similarly to \cite{xiao2022}, Bradford et al.~\cite{bradford2018} maximize an acquisition function to select points to visit, and then use a TSP solver.
In \cite{guestrin2005}, Guestrin et al. use similar Bayesian optimization methods to nearly optimally place sensors in a region.
While these approaches maximize a stronger information-theoretic objective, they do not allow for or incorporate operational constraints like travel budget.
This is further elaborated in~\cite{marchant2014, binney2010}.
Moreover, their offline nature does not incorporate new data as it becomes available, or make planning online feasible.

In order to make plans adaptive to new information, path planners often utilize receding horizon optimization.
While there is no literature for receding horizon GP path planning, GPs have been used in this fashion in the Model Predictive Control (MPC) literature.
In~\cite{polczEfficientImplementationGaussian2023,maiwormOnlineLearningbasedModel2021}, MPC is used to obtain control inputs robust to model uncertainty described by the GP.
As such, GPs are used to inform system dynamics, rather than motivate exploration objectives.
Moreover, the fact that additional information will be gained at each time step is not explicitly included in the optimization problem.

These approaches have two key limitations:
(i) plans are myopic -- they do not anticipate the effects of future information gain, and
(ii) they lack adaptivity, making it difficult to incorporate operational constraints without excessive conservatism.
These limitations motivate our approach.

We present a GP-based receding horizon optimization path planner and explicitly incorporate sensing in the system dynamics.
Unlike existing methods, our approach:
\begin{itemize}
    \item nonmyopically projects the predictive uncertainty multiple steps into the future,
    \item adapts to new data by re-optimizing online, and
    \item naturally incorporates operational constraints.
\end{itemize}
We examine our approach on an algal bloom monitoring scenario and show that it outperforms greedy baselines.
We also compare to a mixed-integer program from~\cite{dutta2025} where the uncertainty is minimized on a fixed graph, and the path is found on a graph.
This differs significantly from our method where the location of the waypoints are selected on an continuous domain subject to a set of dynamic constraints, and minimizes the misclassification probability, instead.
Nonetheless, we include it for completeness.

\subsection*{Motivating Application}
As a motivating application, we study path planning for an autonomous surface vehicle (ASV) equipped with a \textit{chlorophyll a} sensor, which can be used as a biomass proxy to detect algal blooms.
The goal is to generate a plan for the ASV as in the campaigns reported in \cite{oudijk_campaign_2022,Ove23,halvorsen_operational_2025} in which the ASV observes the ocean together with a satellite and a high-fidelity ocean modeling software, SINMOD.
The concept of adaptive ocean sensing and sampling is widely explored, e.g., in \cite{Ler07}, and the use of GP models is common.
For example, \cite{foss_using_2022} and \cite{Fos19} base their plan on SINMOD data for underwater vehicles. 
The benefit of using surface vessels over underwater vehicles, in this context, is that the ASVs operate on the water surface, meaning that getting exogenous data from sources such as satellite is more viable.
In this article, we propose an online look-ahead GP (OLAh-GP) path planning algorithm for an ASV with the goal of identifying the algal blooms in an ocean region with limited sensing budget with high confidence.

\subsection*{Organization}
Following an overview of GPs in \Cref{sec:background}, we present the main method in \Cref{sec:method}. 
We then return to the motivating example in Section \ref{sec:algal}, and test our method using data from a real campaign working to detect algal blooms in a simulated example in \Cref{sec:simulation_setup}.
The results and discussion based on this simulated data are in \Cref{sec:results} and \Cref{sec:discussion}, respectively. 
The setup for the ensuing experimental campaign is laid out in \Cref{sec:experimental-campaign}, before the accompanying results and discussions follow in \Cref{sec:experiments-results} and \Cref{sec:experimental-discussion}.
We offer concluding remarks and future directions in \Cref{sec:conclusion}.

\section{Preliminaries}
\label{sec:background}
\subsection{Notation}
For convenience, ordered sets are denoted by the capital letter of their elements, so $X = (x_{i})_{i=1}^{N}$, and $ \overline{X}$ concatenates the elements of $X$ into a single vector.
Functions with sets as arguments implicitly concatenate the outputs applied to each element: $f(V, x) = [f(v_1, x)^{\transp}\; \cdots\; f(v_N, x)^{\transp}]^{\mathsf{T}}$.
Similarly, $f(X, X)_{ij} = f(x_i, x_j)$.

\subsection{Gaussian Processes}
A process $f: \R^d \to \R$ is a Gaussian process if, for any finite set of points $ x_{1}, \dotsc, x_{n} \in \R^d$, the values of the process at these points follows a multivariate normal distribution~\cite{rasmussenGaussianProcessesMachine2008}.
Specifically, the GP distribution is parametrized by two functions: the mean $m(\cdot): \R^d \to \R$, and the pairwise covariance $k(\cdot, \cdot): \R^d \times \R^d \to \R$, called the kernel.
We write 
\begin{equation}\label{eq:gp-dist}
    f(\cdot) \sim \mathcal{GP}(m(\cdot), k(\cdot, \cdot)), 
\end{equation}
which means that for any $x_1, \dotsc, x_N$, $f$ satisfies
\begin{equation}\label{eq:gp-mvnorm}
    \scalemath{0.85}{
    f(X) \sim 
        \mathcal{N}\left(
            m(X), 
            k(X, X)
        \right).
    }
\end{equation}

A common choice of $k$ \textit{squared exponential} (SE) kernel, 
\begin{equation}\label{eq:covariance}
  k( x_{1}, x_{2}) = \sigma^{2} \exp \left( - \frac{1}{2} ( x_{1} - x_{2})^{\transp}L^{-1} (x_{1} - x_{2}) \right),
\end{equation}
in which $\sigma^{2}$ defines the signal variance, and $L$ defines a characteristic length scale.
Points that are distant according to this scale will be uncorrelated, whereas points close to each other take on similar values.
This choice of kernel stipulates that the underlying process, $f(\cdot)$, is infinitely differentiable.

If we obtain data $y_{i}$ about the process, which has mean value $m(x_{i})$ corrupted by Gaussian noise with zero mean and variance $w^{2}_{i}$, then we can update the posterior distribution of 
$f$ given the data by conditioning the multivariate Gaussian in \eqref{eq:gp-mvnorm}.
Let $\mathcal{D} = \{ (x_i, y_i, w_{i}) \}_{i=1}^N$ be the dataset.
The posterior distribution at a sampling point of interest, $x$, is
\begin{subequations}
    \label{eq:condition}
    \begin{align}
        &f(x) \, | \, \mathcal{D}, x \sim \mathcal{N}(\mu(x \,|\, \mathcal{D}), \sigma^2(x \,|\, \mathcal{D} )) \\ 
        &\mu(x\,|\, \mathcal{D}) = m(x) \nonumber \\ 
        & \qquad + k(x, X) \left[k(X, X) + \Sigma\right]^{-1} (\overline{Y} - m(X)) \\
        & \sigma^2(x\,|\, \mathcal{D}) = k(x, x) \nonumber \\ 
        & \qquad - k(x, X) \left[k(X, X) + \Sigma\right]^{-1} k(X, x),\label{eq:condition_variance}
    \end{align}
\end{subequations}
where $\Sigma = \mathrm{diag}(w_{1}^{2}, \dotsc, w_{n}^{2})$.

\section{Look-ahead Gaussian Process Path Planning}
\label{sec:method}

To react to incoming data and enable online learning for the objective function, we propose the following approach:
First, we model the underlying variable of interest as a GP with prior mean $m(\cdot)$ and fixed kernel $k(\cdot, \cdot)$. 
Then, we employ a receding horizon optimization problem whose decision variables are the waypoints for the path.

The approach is presented in the Online Look-Ahead GP Planning (OLAh-GP) algorithm in \Cref{alg:mpc_alg}.
The idea is for data from any source to be integrated into the agent's model, and for the agent to produce a new plan if there is enough new data.
Afterwards, we propose a form for the optimal control problem which is to be solved in real-time to determine future waypoints.
We assume that we have access to a sufficient amount of initial data.

\begin{algorithm2e}[h]
\caption{OLAh-GP Path Planning}
    \KwData{Prior data set $\mathcal{D}_{\mathrm{prior}}$ \\ 
    Optimization horizon $N$ \\ 
    GP kernel $k(\cdot, \cdot)$ \\ 
    Initial position $r(0)$
    }
    \KwResult{Final collected dataset $\mathcal{D}$}
    Initialize $\mathcal{D} \gets \emptyset$\;
    Initialize $m(x_{i}) \gets y_{i}$ for $(x_{i}, y_{i}, 1)$ in $\mathcal{D}_{\text{prior}}$\;
    $\mu(x) \gets m(x)$ for $x \in \mathcal{X}$\; 
    $\sigma(x) \gets k(x, x)$\;
    $r_\mathrm{target} \gets r(0)$\;
    \SetAlgoVlined
\While{the agent is in operation}{
    \If{at a waypoint} {
        Acquire a measurement $(x_{t}, y_{t}, w_{t})$\;
        $\mathcal{D} \gets \mathcal{D} \cup (x_{t}, y_{t}, w_{t})$\;
        $\mu(x), \sigma(x) \gets GPupdate(m, k, \mathcal{D})$ as in \eqref{eq:condition}\;
        $(r_1, \dotsc, r_N) \gets $ solution to \eqref{eq:mpc}\;
        $r_{\mathrm{target}} \gets r_{1}$\;
    }
    Move towards $r_{\mathrm{target}}$\;
}
\Return $\mathcal{D}$\;
\label{alg:mpc_alg}
\end{algorithm2e}
\vspace*{-2mm}
We now present the optimal control problem to be solved online for a specified future horizon.
\begin{subequations}
\label{eq:mpc}
    \begin{align}
        R^{*} = \argmin_{R \in \Omega^{N}} \quad & \phi_t\left(\mu(\cdot \,| \mathcal{D}_{t}, R), \sigma^{2}(\cdot \,| \mathcal{D}_{t}, R), R, r_0\right) \\ 
        \mathrm{s.t.} \quad & h( R) \leq 0 \\ 
        & r_{0} = r(0),
    \end{align}
\end{subequations}
where $r(0)$ is the current position of the agent at the time the optimization problem is solved and $r_0$ is the initial value for the position for the optimal control problem.
Here, $\mathcal{D}_{t}$ denotes the data set at time step $t$, $h(\cdot)$ is an arbitrary inequality constraint, and $\phi_t(\cdot, \cdot)$ is the cost function of interest at time-step $t$.
The decision variable, $ R = (r_{1}, \dotsc, r_{N})$, is the set of $N$ waypoints, and $R^*$ is the decision variable that optimizes the cost function.

Importantly, we care about costs which depend on the future information state of the agent. 
Our main insight is to model the information state with GPs, in which case, in expectation, we have that 
$\mathbb{E}_{ f(R)}\left[\mu(x \, | \mathcal{D}_{t}, R)\right] = \mu(x\, | \mathcal{D}_{t})$, and we have a closed-form expression for 
$\sigma^{2}(\cdot \, | \mathcal{D}_{t}, R)$.
The most expensive computations in \eqref{eq:mpc} are the calculations of this closed form expression.
In particular, the decision variables appear in a matrix inverse:
\begin{equation}\label{eq:sigup}
\scalemath{0.92}{
    \begin{aligned}
    &\sigma^{2}(x\,|\, \mathcal{D}_{t}, R) = k(x, x) \\ 
    &\qquad - \begin{bmatrix}
        k(X_t, x) \\
        k(R,x) \\
    \end{bmatrix}^{\mathsf{T}} \begin{bmatrix}
        k(X_t, X_t) & k( x, R) \\
        k( R, x) & k( R, R) \\
    \end{bmatrix}^{-1} \begin{bmatrix}
        k(X_t,x) \\
        k(R,x) \\
    \end{bmatrix}.
    \end{aligned}
    }
\end{equation} 
Solving an optimization problem in terms of $\sigma^{2}$ is expensive as $\mathcal{D}$ gets more data over the course of the experiment.
Reference \cite{krauseMoreEfficientRankone2015} shows how to add a single data point to an existing regression in $O(|\mathcal{D}_{t}|^2)$ time.
It is well known that an incremental update for $N$ new data points is on the order of $O(|\mathcal{D}_{t}|^2 N^3)$.
This may be suitable for very small $N$, but for increased scalability of the optimization problem \eqref{eq:mpc} we propose using an approximation that decouples existing data and optimization variables.
The Bayesian Committee Machine (BCM) approximation presented in \cite{trespBayesianCommitteeMachine2000} achieves this decoupling while sacrificing little accuracy in the variance.
While Sparse Variational methods (SVGP) \cite{titsias2009} are well studied, they do not decouple the problem, and instead cast an additional optimization over inducing points.
Compressing the existing data using SVGP methods is a topic for future work.
Comparisons with other approximations are found in \cite{pmlr-v37-deisenroth15}.
This yields the posterior variance 
\begin{equation}\label{eq:sigapprox}
    \frac{1}{ \sigma^{2}(x \,|\, \mathcal{D}_{t}, R) } \approx \frac{1}{\sigma^2(x \,|\, \mathcal{D}_t)} + \frac{1}{\sigma^{2}(x \,|\, R)} - 1,
\end{equation}
where $\sigma^{2}(x \,|\, \mathcal{D}_t)$ is as in \eqref{eq:condition_variance}, and $\sigma^{2}(x\,|\, R)$ is as in \eqref{eq:condition_variance} but with only the data points over $ R$.

\section{Monitoring of Algal Blooms}
\label{sec:algal}
Algae blooms are marine phenomena in which algae rapidly proliferate in an ocean region.
These blooms can be greatly destructive to coastal aquaculture and are of interest to biologists to monitor ocean ecosystems.
\textit{Chlorophyll a} concentration can be used as an indicator for certain types of algal blooms \cite{stumpf_monitoring_2003}.
Our objective is to identify algal blooms in an ocean region.
We model an underlying function, a map of ocean surface chlorophyll concentrations $f$, using a GP with SE kernel with known length scale $L = \ell^2 I_2$, where $I_2$ is the $2\times2$ identity matrix.
Our problem setup consists of initial data from a simulator as our prior mean for the GP and 
an autonomous surface vehicle which can acquire accurate in-situ measurements.

We consider a simplified model for classification of algal blooms in which when $f(x)$ is greater than some threshold $\gamma$, $x$ contains a bloom.
Using the fact that $f(x) \,|\, \mathcal{D}, x \sim \mathcal{N}(\mu(x), \sigma^{2}(x))$ from \eqref{eq:gp-mvnorm}, we can obtain a misclassification probability, $p(x)$, at every point.
The probability of misclassifying a point is given by 
\begin{equation}\label{eq:erf}
\scalemath{0.95}{
\begin{split}
    &p(x\, | \, \mathcal{D}_{t},R) = \Pr[f(x) \geq \gamma \,|\, \mu(x \,|\, \mathcal{D}_t) \leq \gamma] \\
    &\qquad= \Pr[f(x) \leq \gamma \,|\, \mu(x \,|\, \mathcal{D}_t) \geq \gamma] \\ 
    &\qquad = \int_{-\infty}^{\gamma} \mathcal{N}(f; \mu(x \,|\, \mathcal{D}_t), \sigma^2(x \,|\, \mathcal{D}_t, R)) \dd f \\
    &\qquad= \Phi\left(-\frac{|\mu(x) - \gamma|}{\sigma(x\,|\, \mathcal{D}_{t}, R)}\right),
\end{split}}
\end{equation}
where $\mu, \sigma$ are as in \eqref{eq:condition} based on the current data $\mathcal{D}_t$ and the path $R$.
$\Phi(z)$ is the cumulative normal distribution.
This is a common approach, employed by \cite{ foss_using_2022, berget2022, ge2022}, with slight variations in the objective.

With this, we can specify the optimal control problem in \eqref{eq:mpc} for our use-case:
\begin{subequations}
\label{eq:mpc2}
    \begin{align}
        \argmin_{R \in \Omega^N} &  \sum_{x\in \mathcal{X}}  p(x \, | \, \mathcal{D}_{t}, R) + \lambda_1 \sum_{i=0}^{N-1} \norm{r_{i+1}-r_i}_2 &&\label{eq:OCP_waypoints_simple}\\
      \mathrm{s.t.} \, & \sum_{i=0}^{N-1}  \norm{r_{i+1}-r_i}_2 \leq L_{\mathrm{max}}, && \\
      \, & r_{0} =   r(0), &&
    \end{align}
\end{subequations}
where $L_{\mathrm{max}}$ is the maximum distance of the total path.

The goal here is to minimize the overall misclassification probability over the domain $\mathcal{X}$ which is a 2D grid of uniformly spaced points.
The second term (\ref{eq:OCP_waypoints_simple}) regularizes the trajectory length, and orders the waypoints efficiently. 
Finding the ordering of waypoints with minimum path length is an instance of the traveling salesman problem, which is NP-hard and so we provide no guarantees that this finds the most efficient ordering of waypoints.
However, in our experiments, generated paths were empirically in optimal order.
Additionally, the starting point $r_0$ is included as a constraint in the optimization problem.
This objective makes distance a foreboding objective in this formulation, as opposed to an afterthought.
$\lambda_1$ is chosen to compete with the main objective and favors shorter paths.
This means that over a single horizon, the agent ensures that it makes some progress towards the objective.
This is a desirable property since, if the mission ends unexpectedly, the agent still has made meaningful progress.
Having some greediness in a medium horizon makes paths optimal in the short-to-medium term, and is similar to the \textit{anytime property} discussed in \cite{manjanna2017}.
Finally, the (9b) enforces a maximum length, $L_\mathrm{max}$ on the short-horizon plan.

To maintain a tractable estimate of the \textit{chlorophyll a} distribution $f$ over the domain, we discretize the domain over a grid of \verb+RES+$\times$\verb+RES+ uniformly spaced points, $\{x_i\}_{i=1}^{RES^2} = \mathcal{X}$.
$\Omega = [-s / 2, s/2]^2$ is the exploration region, where $s$ is the side length of the region.
The prior data and satellite images are compressed by keeping the sample nearest to each grid point, though other compression techniques can be used.
We argue that the coarse discretization is not restrictive as long as the width of each cell is $ \ll \ell$, which limits aliasing and suffices for the optimization objective. 
For our experiments, we replan every time the surface vessel takes a measurement upon reaching a waypoint.

\subsection*{Computational Complexity} 
We briefly discuss the computational complexity of our approach. 
The cost of $GPupdate$ is $O(|\mathcal{D}_t|^2)$, which is the cost of performing a rank-one update of the Cholesky decomposition representing the state of the GP.
There is a further $O($\verb+RES+${}^2|\mathcal{D}|_t^2)$ cost to computing the posterior mean and variance from the matrix multiplications in \eqref{eq:condition}.
The actual misclassification probability sum is done in time $O($\verb+RES+${}^2)$. 
The optimization problem is then cast on the posterior using the approximation in \eqref{eq:sigapprox} which has complexity $O(N^3 + N^2$\verb+RES+${}^2)$ for symbolically computing the Cholesky decomposition and subsequent posterior variance and misclassification probability over $\mathcal{X}$.
Moreover, we note that we can periodically reset the prior by incorporating data measured thus far into the prior to avoid $|\mathcal{D}_t|$ growing unbounded. 
We provide wall-clock runtimes in \Cref{sec:results}.

\subsection*{Baseline Comparisons}
We compare our approach to three baseline implementations.
Few methods incorporate our cost approach with GP exploration.
Most current work employing GP exploration uses a greedy approach to selecting the next measurement, such as \cite{binney2010, manjannaHeterogeneousMultiRobotSystem2018}.
However, they directly use the posterior variance, and not a function of the variance.
We mimic these greedy approaches with the cost presented in \eqref{eq:mpc2} by setting the horizon of our approach to $N=1$.
We call this the ``greedy baseline''.
The second baseline implementation imitates the approach in \cite{xiao2022} where a single plan is computed before the start of the mission. 
In this case, we optimize \eqref{eq:mpc2} a single time with $N=20$ with the prior dataset and execute the resulting path without incorporating new data.
We call this the offline, or ``static'', baseline.
Finally, we compare to the Mixed Integer Program (MIP) approach from~\cite{dutta2025}.
This approach optimizes the posterior variance by casting the problem as a convex program on a graph.
Our objective is not convex, so we see how well optimizing the posterior variance fares in our approach with this baseline. 
Due to the poor scaling of the MIP, we restrict the graph to a grid of $8 \times 8$ evenly spaced and fully connected nodes on the domain.
Note that since this approach plans on a graph, the path length constraint is enforced by limiting the number of visited nodes rather than an explicit path length contstraint.

\section{Simulation Setup}
\label{sec:simulation_setup}

The simulation setup closely follows the field experimental setup in \cite{halvorsen_operational_2025} for algal bloom detection.
The assets for this paper comprise a high-fidelity ocean modeling software, SINMOD, and the aforementioned ASV.
These provide: a) a high quality prior for our GP, and b) autonomous, high precision, in-situ measurements, respectively.
The goal of this experiment is for the ASV to use the prior from SINMOD in conjunction with its own measurements to correctly identify algal blooms.

The ASV starts only with prior data.
As it collects sensor data, it updates its belief of the overall \textit{chlorophyll a} concentrations and replans to incorporate this new information.
The ASV moves relatively slowly, and so it only explores a small subset of the map. 

\begin{figure}[!htbp]
    \centering
    \legendline{color=GreenYellow}{$m(x) = \gamma$ contours} \quad
    \includegraphics[width=0.615\linewidth, keepaspectratio]{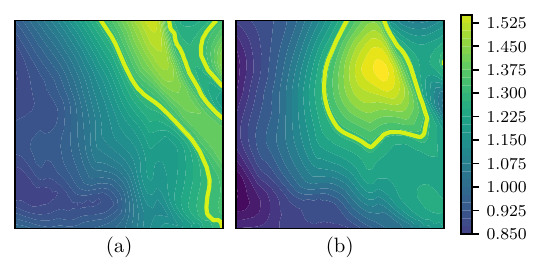} 
    \includegraphics[width=1.2in, keepaspectratio]{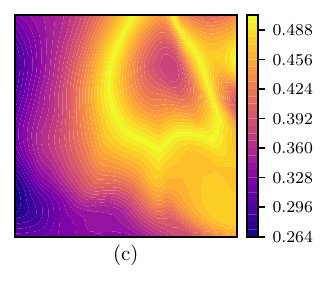} \\ 
    \vspace*{-4mm}
    \caption{\textit{Chlorophyll a} values (mg m$^{-3}$) used in simulation. Panel (a) contains SINMOD \textit{chlorophyll a} predictions from September $3^\mathrm{rd}$, 2025 taken to be ground truth.
        Panel (b) shows the ground truth in (a) corrupted by correlated GP noise taken to be an artificial ``prior''.
        The yellow curves show the decision threshold $\gamma$ contours.
        (c) contains the initial misclassification probability, \eqref{eq:erf}, of (b), i.e. the initial cost seen by the ASV.
    }
    \label{fig:information_2x2}
\end{figure}

SINMOD is a hydrostatic z-level physical-biogeochemical ocean model \cite{slagstad_modeling_2005, wassmann_modelling_2006}.
In this study, the model is run daily to produce short-term forecasts.
The model is run in a 4-step nested setup consisting of domains with resolutions ranging from 20 km down to 160 m.
The higher resolution domains, used in our setup, use atmospheric forcing from the MEPS dataset\footnote{\url{https://www.met.no/en/projects/metcoop}}. 
Phytoplankton is nudged using chlorophyll estimates from the Global Ocean Colour L4 from Satellite Observations \cite{acri_france_global_nodate} in the 800 m domain and Sentinel 3's Ocean and Land Colour Instrument \footnote{\url{https://user.eumetsat.int/catalogue/EO:EUM:DAT:0178}} in the 800 m domain.

\Cref{fig:information_2x2} shows the \textit{chlorophyll a} concentration estimations from SINMOD used for our experiments and the corresponding misclassification probabilities. 
For the remainder of the paper, all heatmaps are over the $s\times s$ exploration region.
We take SINMOD data to be the ground truth, and our simulated prior is the ground truth corrupted by GP noise.
In an operational setting, the prior would be the SINMOD data itself, and we would have no access to ground truth.
We use this ground truth here to get a quantitative sense for the performance of our algorithm.
In \Cref{sec:experimental-campaign}, we will use our approach without access to ground truth.

The map we use is a small subset of the available SINMOD data, whereas we use a larger set of the data to find the $\ell$ of best fit for the kernel.
In order for our prior to have flaws, we simulate correlated noise by drawing a random sample from the distribution $\mathcal{GP}
        \left(
            0, 
            0.2 \exp \left[ 
                -\frac{1}{2} \frac{\norm{x_1 - x_2}^2}{(s/8)^2}
            \right]
        \right),
$
where $s = \qty{4800}{\metre}$ is the size of the map, and we ensure that the prior is non-negative by making the added noise non-negative.
The constants used in the optimal control problem are presented in \Cref{tab:setup-constants}.
\vspace*{-1mm}
\begin{table}[!htbp]
\caption{Optimization constants}
\vspace*{-1mm}
\centering
  \begin{tabular}{ c l l l }
 \toprule
 Variable & Description & Value & Units \\
 \midrule
 \verb+N_POINTS+ & Map samples grid size & 128 & - \\
 \verb+RES+ & Optimization grid size & 32 & -  \\
 \verb+N_STEPS+ & Total experiment length & \nsteps & - \\
 $N$ & Horizon length & \horizon & -  \\
 $L_{\mathrm{max}}$ & Maximum distance & \lmax & \unit{m} \\
 $\ell$ & GP length scale & \lscale & \unit{\m} \\
 $\lambda_1$ & Distance regularization  & \lda & \unit{\per\m} \\
 $\sigma$ & GP Variance & $1.0$ & \unit{\mg \per \m^3} \\
 $r(0)$ & Starting point & $\mathbf{0}$ & \unit{m} \\
 \bottomrule
\end{tabular}
  \label{tab:setup-constants}
\end{table}
The optimal control problem in \eqref{eq:mpc} is solved using CasADi~\cite{anderssonCasADiSoftwareFramework2019} and IPOPT~\cite{wachterImplementationInteriorpointFilter2006}.
The solver is hot-started with solutions from the optimization at the previous waypoint.
The initial guess for the first iteration is a regular n-gon centered at the origin with radius $s / 4$. The choice of $L_{\mathrm{max}}$ at \lmax \ m gives a maximum  expected time span of the planning horizon between 1.5 and 3.5 hours with a speed over ground between 0.5 and 1.2 m/s, giving an average waypoint spacing of at least 17 minutes if the optimizer activates the constraint. 

\section{Results}
\label{sec:results}

We now present simulation results through a series of figures at different stages of the exploration task.
Our goal with these figures is to illustrate the behavior of the system as new data emerges, and show that planning is nonmyopic.

\begin{figure}[!ht]
    \centering
    \legendline{color=NavyBlue}{OLAh-GP} \quad
    \legendline{color=Red}{MIP} \quad
    \legendline{color=orange}{Greedy} \quad \\
    \legendline{color=Green}{Static} \quad
    \legenddoublecircle{white}{white}{Starting position} \quad
    \includegraphics[width=0.9\linewidth]{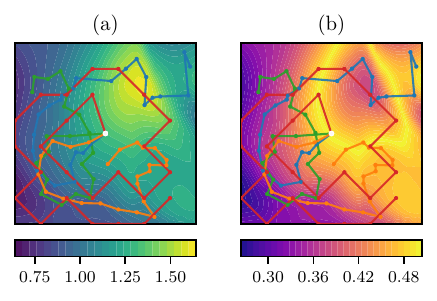}
    \vspace*{-4mm}
    \caption{The path of the ASV for different algorithms over (a) the prior and (b) initial misclassification probabilities. }
    \label{fig:total_paths}
\end{figure}

\begin{figure}[!ht]
    \centering
    \vspace*{-5mm}
    \legendline{color=MidnightBlue}{Executed path} \quad
    \legendline{color=orange}{Planned path} \quad \\
    \legenddoublecircle{black}{LimeGreen}{Starting position} \quad
    \legenddoublecircle{white}{LimeGreen}{Current position} \quad
    \includegraphics[width=0.95\linewidth]{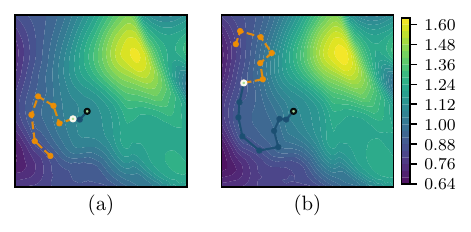}
    \vspace*{-4mm}
    \caption{The path of the ASV over its current \textit{chlorophyll a} estimates (mg m$^{-3}$). 
        (a) shows the position of the ASV after 2 time steps, and (b) after 9 timesteps.
        The dotted orange line shows the current plan of the ASV, i.e. the optimal solution it found, while the solid blue line shows the executed trajectory up to that point.}
    \label{fig:path_with_prediction}
\end{figure}

\begin{figure}[!h]
    \centering
    \vspace*{-5mm}
    \includegraphics[width=0.8\linewidth, keepaspectratio]{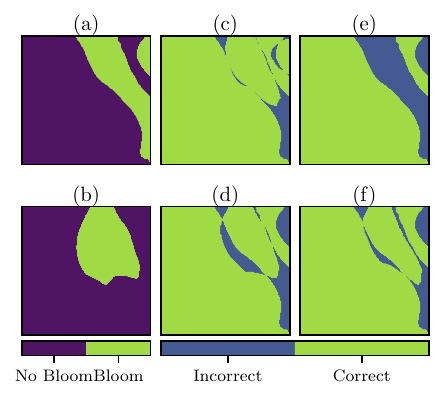}
    \vspace*{-4mm}
    \caption{Subfigures (a) and (b) show the classifications of the region based on the ground truth and prior, respectively.
    Subfigures (c), (d), and (e), (f) show the error in classification for our approach, the greedy baseline, static plan, and MIP baseline, respectively.}
    \label{fig:misclassification}
\end{figure}

\begin{figure}[!ht]
    \vspace*{-2mm}
    \centering
    \legendline{color=NavyBlue}{OLAh-GP} \quad
    \legendline{color=Dandelion}{MIP} \quad
    \legendline{color=orange}{Greedy} \quad
    \legendline{color=Green}{Offline} \quad
    \includegraphics[width=0.95\linewidth, keepaspectratio]{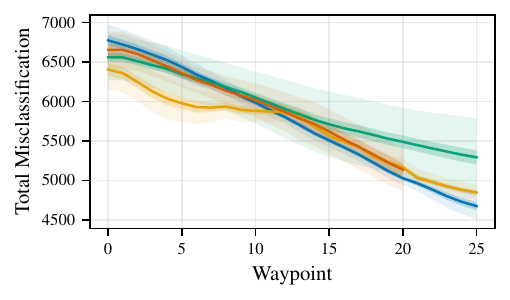}
    \vspace*{-4mm}
    \caption{Comparison against baselines. The shaded regions indicate $\pm1$ standard deviations from the mean, from 10 experiments for each approach.
    The darker shaded regions indicate $\pm1$ standard error in the mean over the 10 experiments, they are barely visible for our planner and the MIP planner.
    30 experiments were used for the greedy approach.}
    \label{fig:baselines}
\end{figure}

In Fig.~\ref{fig:total_paths}, the paths for the three different path generation schemes are shown.
All plans start at the same point, but quickly diverge.
The executed path by the OLAh-GP algorithm is shown in Fig.~\ref{fig:path_with_prediction} at two different stages of the task, showing how the planned path differs from the executed path even over a short horizon.
This highlights the adaptive nature of the algorithm to measurements collected online, and the need for adaptivity.
Fig.~\ref{fig:misclassification} shows the performance in correctly identifying algal blooms.
As the figure shows, the OLAh-GP planner classifies the largest area correctly, followed by the MIP planner, greedy approach and the offline OLAh-GP path.
The evolution of the total costs is compared in Fig.~\ref{fig:baselines}. 
For all comparisons, the total cost is the sum of misclassification probabilities over the \verb+N_POINTS+$\times$\verb+N_POINTS+ grid, i.e. \eqref{eq:mpc2} with $\lambda_1 =0$.
Initially, the costs are similar but over time, OLAh-GPs medium-term planning and adaptability allow it to surpass other methods.
We also note that our method was markedly more consistent in reducing error than the other approaches, as indicated by the shaded regions.

Summary statistics of runtime for our method and baselines are shown in \Cref{tab:IPOPT-output}.
After each measurement, the agent replans for the OLAh-GP and greedy algorithm, but not for the offline algorithm. 
For our simulated experiments, we run 10 experiments each with 25 waypoints, except for the offline algorithm, where memory usage limited the maximum horizon length.
For improved readability, we omitted the results of running our method with an optimization horizon of 7 steps. 
We do report the computation time to give a sense of the computational scaling of the approach with horizon.
Since we use the SINMOD data as a prior, we incur no cost to updating the model to a high-density data initial estimate and therefore do not report the associated computational cost.

\begin{table}[!htbp]
\caption{Optimization statistics}
\centering
\setlength{\tabcolsep}{4pt}
\begin{tabular}{ L{1.23cm} c c c c}
\toprule
 \textbf{Method} & \makecell{$GPupdate$ \\ time (\unit{\ms})} & \makecell{Setup \\ time (\unit{\s})} & \makecell{Optimization \\ time (\unit{\s})} & \makecell{Iterations} \\
 \midrule
 {\scriptsize OLAh-GP 5} & $50 \pm 32$ & $3.7 \pm 0.1$ & $0.99 \pm 0.70$ & $37 \pm 25$ \\ 
 {\scriptsize OLAh-GP 7} & $51 \pm 33$ & $6.8 \pm 0.1$ & $3.4 \pm 5.9$ & $56 \pm 94$ \\
 \midrule
 Greedy& $57 \pm 34$ & $0.62 \pm 0.04$ & $21 \pm \SI{7}{ms}$ & $16 \pm 5$ \\
  {\scriptsize MIP} & $50 \pm 33$ & $ 7 \pm \SI{5}{ms}$ & $17 \pm 9$ & $8.6\text{k} \pm 3.9\text{k}$ \\
 Offline & N/A & $81 \pm 5$ & $730 \pm 1200 $ & $670 \pm 1100$ \\
\bottomrule
\end{tabular}
  \label{tab:IPOPT-output}
\end{table}

\section{Discussion}
\label{sec:discussion}

These results show several advantages to the proposed approach. 
Both Fig.~\ref{fig:misclassification} and Fig.~\ref{fig:baselines} show that planning for the medium term offers significant advantages.
They also show clearly that reacting to new data is essential in completing mission objectives.
However, the total cost curves Fig.~\ref{fig:baselines} differ little between the offline, MIP, and OLAH-GP approaches. 
In our experiments, the prior is usually in near conformity with the ground truth.
As a result, the online measurement will not change the belief of the world drastically.
Despite this, data collected online significantly changes the path planning, which can be seen by the differences in the offline and the online OLAh-GP maps in Fig.~\ref{fig:total_paths}.
It also greatly affects the regions that are correctly and incorrectly classified in Fig.~\ref{fig:misclassification}.


\Cref{tab:IPOPT-output} demonstrates that the nonlinear solver finds solutions quickly.
This makes our approach feasible for real-time applications.
In particular, the application described here has large distances between waypoints relative to the speed of the ASV, and so the time to find solutions is negligible compared to overall experiment time.

The optimal control problem shown in \eqref{eq:mpc} is both non-convex and non-linear, so we have no guarantees of reaching the global minima.
We do not provide proofs of convergence to global optima, but the results indicate that the local minima that we do find give reasonable results.
As can be seen in Fig.~\ref{fig:baselines}, OLAh-GP outperforms the baseline algorithms over a longer timeline, as expected.
Moreover, on all our experiments, we did not encounter infeasible regions of the search space.
However, the reader should be aware that we are hot-starting the nonlinear solver with the solution obtained at the previous waypoint.
This procedure makes it more likely for the optimizer to sequentially return similar local minima.

\section{Experimental Campaign}
\label{sec:experimental-campaign}

We tested our method using a primarily wave-propelled ASV in a coastal location with strong tides and wind.
The waypoints were calculated on a computer onshore and sent directly to the ASV using the LSTS toolchain \cite{pinto2013lsts}.
The lower-level control structure that allows the ASV to follow waypoints is described in \cite{Dal22,Dal25}. 

Our experimental ASV had limited propulsion, so we amended the cost function to account for additional operational constraints.
The ASV needs waves to generate propulsion force, so it struggles when the current gets too strong, and is affected by the current.
We introduce a cost on the waypoints whenever the predicted current (from SINMOD) exceeds a pre-specified threshold.
To account for wind, we add a cost based on the direction and magnitude of the wind, and penalize moving upwind less more than we reward moving downwind.
Modeling the effects of waves is beyond the scope of this work.
The new cost function, an updated version of \eqref{eq:OCP_waypoints_simple}, is now
\begin{align}
\label{eq:updated_cost}
    \begin{split}
        R^{*} &= \argmin_{R \in \Omega^N} \quad \sum_{x\in \mathcal{X}} p(x \, | \, \mathcal{D}_{t}, R) \\ 
        &+ \lambda_1 \sum_{i=0}^{N-1} \norm{r_{i+1}-r_i}_2 + \lambda_2 c(R) +\lambda_3 w(R),
    \end{split}
\end{align}
where the first line is identical to that of \eqref{eq:OCP_waypoints_simple}. 
The other regularization coefficients, $\lambda_2$ and $\lambda_3$, are also positive constants set to $10^{-8}$ and $10^{-5}$, respectively. 
$c(R)$ is the cost associated with the current, and $w(R)$ is the cost associated with the wind.
$c(R)$ is defined as 
\begin{equation}
    c(R) = \sum_{i=0}^{N-1}\begin{cases}
        \norm{u_{i+1}} \norm{r_{i+1}-r_i}^2, & \text{if } \norm{u_i} \ge \epsilon \\
    0,  & \text{otherwise}
\end{cases}
\end{equation}
where $u_i$ is the velocity of the current at waypoint $r_i$.
If the magnitude of the current at the waypoint is greater than $\epsilon$, we factor this into the objective.
Otherwise, no cost is added. 
The wind cost $w(R)$ is defined as 
\begin{equation}
    w(R) = -\sum_{i=0}^{N-1} \mathrm{LeakyReLU}\left[(r_{i+1}-r_{i})^{\transp} W\right],
\end{equation}
where the wind vector $W$ is the average wind vector based on the weather prediction retrieved from the local governmental wind API.
The negative slope of LeakyReLU was set to 0.8.

%
We investigate two scenarios. 
The first follows the simulation setup described in \Cref{sec:simulation_setup} exactly. 
In the second, we take the SINMOD data to be our prior, have no access to ground truth, and use the real in-situ measurements to guide path planning.
For both, we use the real ASV position with the additional costs described above, and compare our approach to the greedy baseline.

\section{Experimental Results}
\label{sec:experiments-results}

\begin{figure}[!ht]
    \centering
    \legendline{color=orange}{OLAh-GP} \quad
    \legendline{color=red}{Greedy} \quad
    \legenddoublecircle{white}{orange}{Starting Position} \quad
    \includegraphics[keepaspectratio]{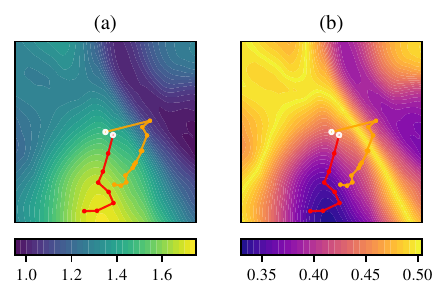}\\
    \vspace*{-3mm}
    \includegraphics[keepaspectratio]{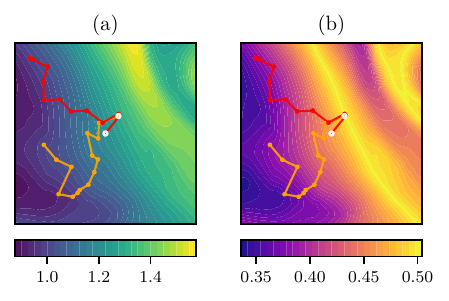}\\
    \vspace*{-4mm}
    \caption{(a) The ASV trajectories for both methods over the prior \textit{chlorophyll a} estimates (\unit{mg/m^3}) for (top) September $3^\mathrm{rd}$, 2025 and (bottom) September $2^\mathrm{nd}$, 2025.
        Panel (b) shows the same over the initial misclassification probabilities.
        The experiments on the top line simulate measurements with SINMOD as ground truth, while the bottom experiments use in-situ measurements as ground truth.
    }
    \label{fig:path_SINMOD_ground_truth}
\end{figure}

\begin{figure}[!ht]
    \vspace*{-2mm}
    \centering
    \legendline{color=NavyBlue}{OLAh-GP} \quad
    \legendline{color=orange}{Greedy} \\
    
    \includegraphics[width=0.9\linewidth, keepaspectratio]{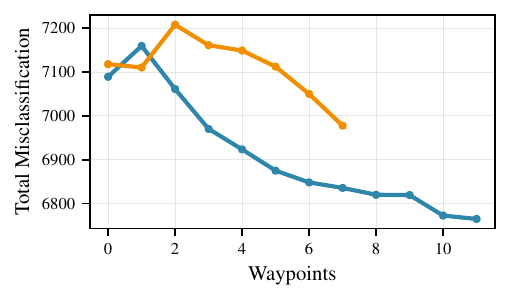} \\
    \vspace{-2mm}
    \includegraphics[width=0.9\linewidth, keepaspectratio]{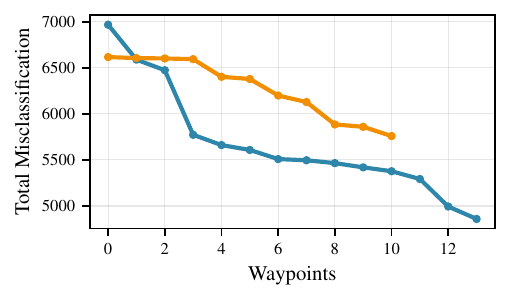} \\ 
    \vspace*{-4mm}
    \caption{Top: comparison of our method against greedy baseline using simulated measurements from SINMOD as a function of the number of waypoints reached.
    Bottom: the same for in-situ measurements.}
    \label{fig:baselines_SINMOD_ground_truth}
\end{figure}

Figs.~\ref{fig:path_SINMOD_ground_truth} show the experimental paths computed by the OLAh-GP planner and the greedy baseline for either simulated measurements or in-situ measurements.
Comparison of the objective value as a function of waypoint completion for both methods are shown in Fig.~\ref{fig:baselines_SINMOD_ground_truth}.

\section{Experimental campaign - discussion}
\label{sec:experimental-discussion}
The main limitation in testing using real measurements is that we cannot hold all environmental variables constant across our experiments.
This means that the baseline is not directly comparable to the designed method as the underlying data cannot be held constant during the experiments. 

However, we still note marked improvement over baseline in both cases.
Our method performs especially well with additional experimental constraints since it makes fewer assumptions about the cost, and plans over a longer horizon.
For example, the greedy planner does not make good use of the wind cost because it has delayed long-term effects.

We note that there are fewer points for the greedy baseline in part because the experiment timeline was constrained.
While the greedy experiments had to be run over a shorter time span, we see that the baseline approach headed towards the corner of the map in the bottom plot of \Cref{fig:baselines_SINMOD_ground_truth}.
We hypothesize that its forward short-term progress would be poor compared to our method, which would strengthen our claims.

Note further that the cost initially increases in the top plot of \Cref{fig:baselines_SINMOD_ground_truth} as the first measurement changed the size of the decision boundary considerably. 
Our method, OLAh-GP, is able to make an efficient plan to account for the new map, while the greedy baseline takes the nearest point with information and moves away from the decision boundary.

Finally, we emphasize that the ASV is capable of taking measurements almost continuously.
In these experiments, we only kept those taken at waypoints, but it may be interesting future work to consider continuous path-planning methods.


\section{CONCLUSION}
\label{sec:conclusion}
The GP-based look-ahead path planning method proposed in this paper plans further into the future than other comparable methods. 
We show that the online look-ahead GP method has benefits over both an offline look-ahead version of the method and a greedy algorithm.
It produces plans that balance accomplishing an objective with operational constraints, achieves the objective with greater efficiency, and adapts to new data.
As such, it has the potential to broadly benefit multi-agent observation systems, should it be complemented with the sensing capabilities of other collaborating agents, be they unmanned aerial vehicles (UAVs), autonomous underwater vehicles (AUVs), or even manual measurements taken by scientists. 

\section*{ACKNOWLEDGEMENTS} 
{As per IEEE guidelines, the authors disclose that Microsoft Copilot, OpenAI's ChatGPT, and Anthropic's Claude were used for code, such as API calls and plotting figures.
}

\bibliographystyle{IEEEtran}
\bibliography{finalReferences} 

\end{document}